\documentclass[10pt,twocolumn,letterpaper]{article}

\usepackage[utf8]{inputenc}
\usepackage[T1]{fontenc}
\usepackage{graphicx}
\usepackage{longtable}
\usepackage{wrapfig}
\usepackage{rotating}
\usepackage[normalem]{ulem}
\usepackage{amsmath}
\usepackage{amssymb}
\usepackage{capt-of}
\usepackage{hyperref}
\usepackage[margin=0.85in]{geometry}
\usepackage{newtxtext}
\usepackage{microtype}
\usepackage{flushend}
\usepackage{booktabs,tabularx}
\usepackage{caption}
\usepackage{tikz}
\usetikzlibrary{arrows.meta,positioning}
\usepackage{titlesec}
\titleformat{\section}{\normalfont\large\bfseries}{\thesection}{0.6em}{}
\titleformat{\subsection}{\normalfont\normalsize\bfseries}{\thesubsection}{0.5em}{}
\titlespacing*{\section}{0pt}{1.3ex plus .3ex}{0.65ex plus .2ex}
\titlespacing*{\subsection}{0pt}{1ex plus .2ex}{0.45ex plus .1ex}
\hypersetup{colorlinks=true,urlcolor=blue,citecolor=blue,linkcolor=black}
\author{\texorpdfstring{Ayush Maniar\textsuperscript{1}\quad Eryk Banatt\textsuperscript{1}\\[0.4em]\normalsize Riot Games\\\normalsize\texttt{\{amaniar, ebanatt\}@riotgames.com}\\[0.3em]\small\textsuperscript{1}Equal contribution.}{Ayush Maniar, Eryk Banatt}}
\date{}
\title{Off-the-Shelf Autonomous Agents for Game Playtesting}
\hypersetup{
 pdfauthor={\texorpdfstring{Ayush Maniar\textsuperscript{1}\quad Eryk Banatt\textsuperscript{1}\\[0.4em]\normalsize Riot Games\\\normalsize\texttt{\{amaniar, ebanatt\}@riotgames.com}\\[0.3em]\small\textsuperscript{1}Equal contribution.}{Ayush Maniar, Eryk Banatt}},
 pdftitle={Off-the-Shelf Autonomous Agents for Game Playtesting},
 pdfkeywords={},
 pdfsubject={},
 pdfcreator={},
 pdflang={English}}
\begin{document}

\maketitle
\begin{abstract}
In playtesting games, a human operator is typically required to test most gameplay elements. In this work we present Buddy, an autonomous game solving agent which can solve long time horizon objectives in an unseen 3D role-playing game using off-the-shelf language model systems. Buddy will take actions, receive feedback on those actions, and pursue a goal across many turns. No written walkthrough or model training of any kind is required, and we show that this paradigm holds in a novel environment where the agent must discover solutions via exploration rather than recalling them from pretraining. We demonstrate three downstream capabilities: game solving with no provided walkthrough, logging bug tickets for encountered difficulties, and conversion of successful agent traces into scripted hierarchical task networks for regression testing. Unlike existing playtesting solutions, our framework is fully modular: language models and harnesses can be easily swapped out as capabilities improve over time.
\end{abstract}

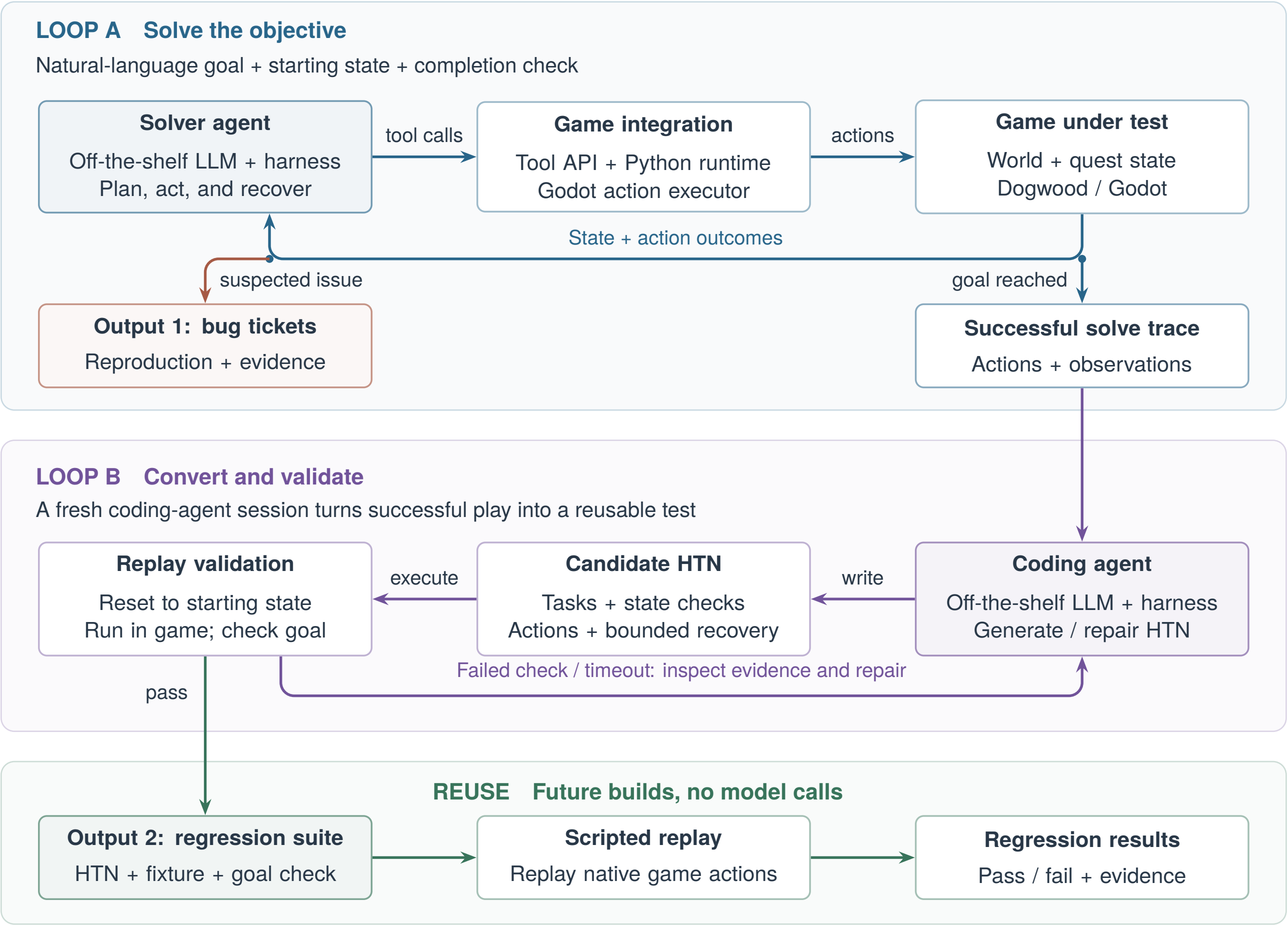
\begin{figure*}[t]
\centering
\begingroup
\definecolor{BuddyInk}{HTML}{283747}
\definecolor{BuddyBlue}{HTML}{28658A}
\definecolor{BuddyPurple}{HTML}{71539A}
\definecolor{BuddyGreen}{HTML}{38745B}
\definecolor{BuddyRed}{HTML}{A65A45}

\begin{tikzpicture}[
  x=1cm,
  y=1cm,
  font=\sffamily\fontsize{8.5}{10.5}\selectfont,
  text=BuddyInk,
  >={Stealth[length=6pt,width=4.5pt]},
  panel/.style={rounded corners=5pt, line width=0.5pt},
  box/.style={
    rounded corners=3pt,
    line width=0.65pt,
    fill=white,
    align=center,
    text width=4.05cm,
    minimum height=1.45cm,
    inner xsep=5pt,
    inner ysep=5pt
  },
  artifact/.style={box, minimum height=1.10cm},
  flow/.style={->, line width=1pt, rounded corners=5pt},
  feedback/.style={flow, draw=BuddyPurple},
  edge label/.style={
    font=\sffamily\footnotesize,
    align=center,
    inner sep=2pt
  },
  heading/.style={
    anchor=west,
    font=\sffamily\small\bfseries,
    inner sep=0pt
  }
]

\path[panel, draw=BuddyBlue!25, fill=BuddyBlue!3]
  (0,0) rectangle (17,-5.40);
\path[panel, draw=BuddyPurple!25, fill=BuddyPurple!3]
  (0,-5.80) rectangle (17,-9.65);
\path[panel, draw=BuddyGreen!25, fill=BuddyGreen!3]
  (0,-10.05) rectangle (17,-12.20);

\node[heading, text=BuddyBlue] at (0.45,-0.40)
  {LOOP A\quad Solve the objective};
\node[anchor=west, inner sep=0pt] at (0.45,-0.86)
  {Natural-language goal + starting state + completion check};

\node[box, draw=BuddyBlue!65, fill=BuddyBlue!7]
  (solver) at (2.70,-2.05) {
    \textbf{Solver agent}\\[4pt]
    Off-the-shelf LLM + harness\\
    Plan, act, and recover
  };
\node[box, draw=BuddyBlue!45]
  (integration) at (8.50,-2.05) {
    \textbf{Game integration}\\[4pt]
    Tool API + Python runtime\\
    Godot action executor
  };
\node[box, draw=BuddyBlue!45]
  (game) at (14.30,-2.05) {
    \textbf{Game under test}\\[4pt]
    World + quest state\\
    Dogwood / Godot
  };

\draw[flow, draw=BuddyBlue]
  (solver.east) -- node[edge label, above=3pt] {tool calls}
  (integration.west);
\draw[flow, draw=BuddyBlue]
  (integration.east) -- node[edge label, above=3pt] {actions}
  (game.west);
\draw[flow, draw=BuddyBlue]
  (game.south) -- (14.30,-3.40)
  -- node[edge label, above=3pt, text=BuddyBlue]
    {State + action outcomes}
  (3.55,-3.40) -- ([xshift=0.85cm]solver.south);
\node[artifact, draw=BuddyRed!65, fill=BuddyRed!5]
  (bugs) at (2.70,-4.55) {
    \textbf{Output 1: bug tickets}\\[3pt]
    Reproduction + evidence
  };
\node[artifact, draw=BuddyBlue!55, fill=white]
  (trace) at (14.30,-4.55) {
    \textbf{Successful solve trace}\\[3pt]
    Actions + observations
  };
\fill[BuddyBlue] (3.55,-3.40) circle[radius=1.5pt];
\fill[BuddyBlue] (14.30,-3.40) circle[radius=1.5pt];
\draw[flow, draw=BuddyRed]
  (3.55,-3.40) -- (2.70,-3.40)
  -- node[edge label, right=3pt] {suspected issue}
  (bugs.north);
\draw[flow, draw=BuddyBlue]
  (14.30,-3.40) -- node[edge label, left=3pt] {goal reached}
  (trace.north);

\node[heading, text=BuddyPurple] at (0.45,-6.28)
  {LOOP B\quad Convert and validate};
\node[anchor=west, inner sep=0pt] at (0.45,-6.74)
  {A fresh coding-agent session turns successful play into a reusable test};

\node[box, draw=BuddyPurple!65, fill=BuddyPurple!7]
  (coder) at (14.30,-7.90) {
    \textbf{Coding agent}\\[4pt]
    Off-the-shelf LLM + harness\\
    Generate / repair HTN
  };
\node[box, draw=BuddyPurple!45]
  (candidate) at (8.50,-7.90) {
    \textbf{Candidate HTN}\\[4pt]
    Tasks + state checks\\
    Actions + bounded recovery
  };
\node[box, draw=BuddyPurple!45]
  (validation) at (2.70,-7.90) {
    \textbf{Replay validation}\\[4pt]
    Reset to starting state\\
    Run in game; check goal
  };

\draw[flow, draw=BuddyPurple] (trace.south) -- (coder.north);
\draw[flow, draw=BuddyPurple]
  (coder.west) -- node[edge label, above=3pt] {write}
  (candidate.east);
\draw[flow, draw=BuddyPurple]
  (candidate.west) -- node[edge label, above=3pt] {execute}
  (validation.east);
\draw[feedback]
  ([xshift=1cm]validation.south) -- (3.70,-9.18)
  -- node[edge label, above=3pt, text=BuddyPurple]
    {Failed check / timeout: inspect evidence and repair}
  (14.30,-9.18) -- (coder.south);

\node[heading, text=BuddyGreen] at (5.70,-10.47)
  {REUSE\quad Future builds, no model calls};
\node[artifact, draw=BuddyGreen!65, fill=BuddyGreen!7]
  (suite) at (2.70,-11.32) {
    \textbf{Output 2: regression suite}\\[3pt]
    HTN + fixture + goal check
  };
\node[artifact, draw=BuddyGreen!45]
  (replay) at (8.50,-11.32) {
    \textbf{Scripted replay}\\[3pt]
    Replay native game actions
  };
\node[artifact, draw=BuddyGreen!45]
  (report) at (14.30,-11.32) {
    \textbf{Regression results}\\[3pt]
    Pass / fail + evidence
  };

\draw[flow, draw=BuddyGreen]
  (validation.south) -- node[edge label, pos=0.25, left=4pt] {pass}
  (suite.north);
\draw[flow, draw=BuddyGreen] (suite.east) -- (replay.west);
\draw[flow, draw=BuddyGreen] (replay.east) -- (report.west);

\end{tikzpicture}
\endgroup
\caption{
Buddy's two-loop architecture.
\textbf{Loop A} solves a natural-language objective through game actions and feedback, logging suspected bugs along the way.
\textbf{Loop B} converts a successful trace into a hierarchical task network (HTN), replaying and repairing it until validation passes.
The resulting tests run on future builds without model calls.
}
\label{fig:architecture}
\end{figure*}
\section{Introduction}
\label{sec:org4b3e5d8}

Developing a video game requires designers to craft various objectives for a player to interact with. These objectives can be logically correct and properly designed, but still impossible to play via normal means. A required midpoint of the objective could be spawned behind an impassable wall, an NPC could have required dialogue overwritten by another objective, a required object may be impossible to obtain, and so on. Determining if a quest is completable via normal means typically requires time from a human operator, which is expensive and pulls the operator away from other tasks they could be working on.~\cite{titan,planning}

In this work, we present Buddy, an AI game solving agent we built to solve a 3D role-playing sandbox game. Buddy is capable of acting in many ways like a human player: it can observe its surroundings, navigate around the world, speak to NPCs, remember previous locations, and perform many other goal-oriented behaviors required to solve a role-playing game. Buddy is capable of completing the full multi-objective main quest of the game, autonomously operating for 45+ minutes, with no walkthrough or prior knowledge of what to do.

Our main contribution is our practical framework for constructing modular, autonomous game-solving agents. Buddy requires no additional model training, no custom harness, no trained input control policy, no game-specific training dataset, and no walkthrough. The only game-specific integration steps are observation and action tools, a Python runtime, and an in-engine executor. The other elements can easily be swapped out as harnesses and models improve. We design Buddy such that incorporating a similar system into a different, in-development game can easily and quickly be accomplished using strong, inexpensive, off-the-shelf components, for maximum flexibility and performance.

To playtest content packages to verify their completability, the demonstrated autonomous exploration component is not required. Buddy can be provided with a full description of each step of its provided objective (e.g. the game code for a quest) and solve the task without needing to explore. In this work we describe a two-loop system: first, we use Buddy to solve the task in multiple steps using a language model in the loop, in order to demonstrate that the objective can be completed; second, we have Buddy perform a postmortem analysis of its agent trace in order to create a single scripted hierarchical task network which solves the task.~\cite{htn} In doing so, we can scale to thousands of objectives, as the language-model loop becomes a one-time cost, and we can repeatedly test the objective cheaply in the future. 
\section{The Dogwood Sandbox Environment}
\label{sec:orga6ac818}

\begin{figure}[t]
\centering
\includegraphics[width=\columnwidth]{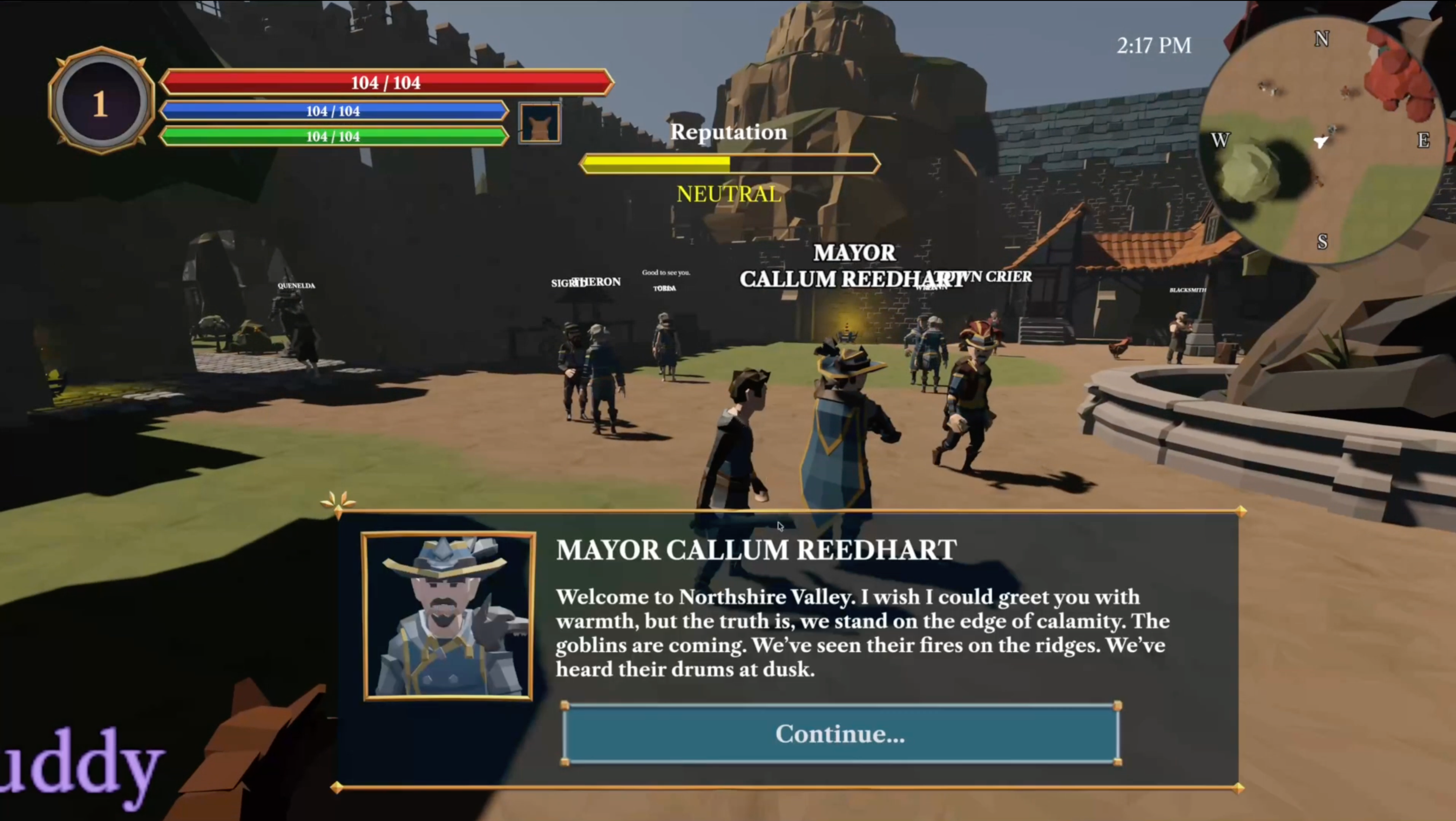}
\caption{\label{fig:org74e0d39}The Dogwood Sandbox Environment.}
\end{figure}

Dogwood is a single-player 3D role-playing game developed in Godot Engine 4.6. This environment is designed to be easy to modify, and includes a main quest with several objectives. Gameplay resembles other traditional fantasy RPGs: the player character can explore the world, speak with NPCs, collect items and gear, fight turn-based battles, and so on. The main quest of Dogwood spans several missions across the environment, testing exploration, fetching items, fighting enemies, and revisiting previously-encountered NPCs and locations. 

We introduce the Dogwood environment for several reasons. First and foremost, as an unpublished research sandbox, it has objectives which are not available for language models to have seen in pretraining. This makes it of substantial interest to game developers, who obviously work inside of unpublished games until their games are publicly released. Existing game-playing evaluations typically use well-established properties like Pokemon Red~\cite{claudepokemon}, Pokemon Showdown~\cite{pokeagent}, RuneScape~\cite{runebench}, Genshin Impact~\cite{lumine}, and Minecraft~\cite{minecollab,voyager}. In all of these examples, rich and detailed instructions for a huge variety of objectives can be found across the internet, making it possible that language models would be able to leverage their existing knowledge in order to make progress solving the game.

Second, we create Dogwood as an example of a game which is well-suited for language model integration. Godot Engine has strong support for script source replacement and reloading, which makes it easy to bolt on a language model game solving agent.~\cite{godotscript} This is still possible in engines like Unreal Engine, but requires script runners to be built to avoid recompiling the game.
\section{Related Work}
\label{sec:org0978f65}

\subsection{From-Gamestate Methods}
\label{sec:org2469120}

In Mindcraft, language models interact with Minecraft via specialized tools which allow them to control their own state and take actions in the game, and MineCollab further extended to collaborative tasks.~\cite{mindcraft,minecollab} Voyager combines an off-the-shelf model with an automatic curriculum, execution feedback, and a growing library of executable skills.~\cite{voyager} SMART uses LLM interpretation of code changes in order to steer reinforcement-learning (RL) agents towards specific gameplay objectives. ~\cite{smart}

Buddy follows a direct line from Mindcraft. We inherit its use of semantic actions, action feedback, remembered places, self-prompting, and code-block generation. Buddy extends this framing to include tools like physical navigation feedback, vision-driven map inspection, and quest and dialogue tools. Many of these capabilities, like execution feedback, code generation,~\cite{codex,claudecode} and even skills management ~\cite{hermes} are now standard features in modern agentic coding frameworks, and can be incorporated in our system through leveraging those harnesses rather than implementing them as a separate game-playing architecture. Buddy applies these capabilities for goal-oriented, developer-directed playtesting. Whereas Voyager builds gradually more complex behavior from a growing library of skills, our trace-to-HTN approach starts from complete solutions, whose shared subplans can later be factored into reusable and composable skills. 
\subsection{From-Pixel Methods}
\label{sec:org933734e}

Lap converts screenshots of a match-three game into structured board state, then uses an LLM to select moves, evaluating coverage and crashes.~\cite{lap} Cradle ~\cite{cradle} uses off-the-shelf foundation models in order to solve games, operating entirely from screenshots and keyboard-and-mouse control; it is better understood as a computer use system that solves games.

In Cradle's follow-up work, Lumine trains a control policy for keyboard-and-mouse actions, which was used to demonstrate long-horizon game solving in Genshin Impact, as well as transfer to new games.~\cite{lumine} SIMA 2 likewise combines a foundation model with embodied training and demonstrates generalization to previously unseen virtual worlds.~\cite{sima2}

In comparison, Buddy primarily operates from privileged game state, and can optionally request screenshots as an option available in its tool corpus. Lumine holds some advantages over Buddy in that it can be used to playtest games with action-driven combat where fast timing and reactions are important concepts to be tested. However, it relies on a specially trained control policy model, compared to ours which can be easily assembled from off-the-shelf components even for unreleased games.
\subsection{Automated QA for Games}
\label{sec:org6d33c87}

A 2022 literature review by Politowski et al. identified workflow integration, explicit testing goals, and test oracles as requirements for automated game testing. ~\cite{politowski} We aim to show progress along these fronts by logging encountered bugs, completing natural-language objectives, and converting to repeatable tests, respectively.

TITAN is the closest to our work, describing similar game state, action guidance, memory, reflection, and bug-hunting in commercial MMOs.~\cite{titan} Existing coding-agent harnesses have also been evaluated in game environments, including DiG-bench and RuneBench, but these are primarily intended to measure capabilities, rather than be used for developer playtesting. ~\cite{digbench,runebench}

Balyo et al. learn a PDDL action model from game logs and use planning to generate test scripts.~\cite{planning} SAGE~\cite{sage} generates regression tests for gray-box game environments using LLM-guided reinforcement learning to train an exploration policy.

Our work differs from TITAN in two main ways. First, we describe a modular framework for repurposing common agent harnesses like Codex~\cite{codex} and Hermes Agent~\cite{hermes} instead of writing a separate, bespoke system. Second, we describe a method to convert these successful agent traces into a repeatable replay artifact, which can be used to cheaply continue testing a solved objective even long after the initial solve. Compared to Balyo et al. and SAGE, our straightforward coding-agent-driven approach allows us to convert the solver trace into a hierarchical task network (HTN) without requiring us to first learn a general planning model of the game or train an exploration policy.
\section{Architecture}
\label{sec:org79a13c3}

Buddy has five main components: 1) an off-the-shelf large language model (e.g. gpt-5.6-luna), 2) an off-the-shelf agentic coding harness (e.g. Hermes Agent), 3) a suite of tools for perceiving and acting in the game (e.g. a tool which will command the player character to walk to a specific location), 4) a Python-side agent runtime (i.e. the server that manages agent sessions and forwards tool calls), and 5) a Godot-side executor (i.e. the game-side code that executes actions and reports outcomes). No model parameter updates are a part of this system at any step.

Figure \ref{fig:architecture} shows the two-loop structure: in loop A, we have the language model agent solve an objective posed in natural language; in loop B, we convert a successful agent trace to a repeatable program which can be run again in the future without language model token spend. There are two primary outputs: bug tickets, and regression testing HTNs.
\subsection{Tool Design for Dogwood}
\label{sec:org06b7116}

Following Mindcraft, we design tools which an off-the-shelf agent can use to observe and interact with the game. ~\cite{mindcraft,minecollab} This lets the agent interact with the game through high-level commands like "talk to this NPC" while relying on the execution layer to handle resolving the target, navigating into range, opening the dialogue, and so on. We similarly adapt Mindcraft's navigation, remembered places, action feedback, and self-prompting.

These tools are designed to expose both perception and action through clear, structured game state, rather than primarily through pixels and keypresses, respectively. In our environment, these cover abilities like local observation, navigation, dialogue, world interaction, inventory management, combat, and so on. Screenshots and programmable map inspection can be used to supplement this structured state, allowing the agent to leverage multimodal capabilities when applicable.

Table \ref{tab:dogwood-tools} summarizes the tool families and their feedback.

\begin{table*}[t]
\centering
\footnotesize
\setlength{\tabcolsep}{3pt}
\renewcommand{\arraystretch}{1.12}
\caption{Tool families used by Buddy for playtesting, and the feedback they return.}
\label{tab:dogwood-tools}
\begin{tabularx}{\textwidth}{@{}>{\raggedright\arraybackslash}p{1.20in} >{\raggedright\arraybackslash}p{3.10in} >{\raggedright\arraybackslash}X@{}}
\toprule
Family & Tools & Main feedback \\
\midrule
State and local observation & \texttt{get\_\allowbreak{}stats}, \texttt{get\_\allowbreak{}nearby\_\allowbreak{}entities}, \texttt{get\_\allowbreak{}nearby\_\allowbreak{}items} & Resources, inventory, quest criteria, local entities/items \\
Navigation and spatial memory & \texttt{go\_\allowbreak{}to}, \texttt{go\_\allowbreak{}to\_\allowbreak{}position}, \texttt{stop}, \texttt{remember\_\allowbreak{}place}, \texttt{list\_\allowbreak{}remembered\_\allowbreak{}places} & Arrival or failure, location, movement feedback, saved waypoints \\
Dialogue & \texttt{talk\_\allowbreak{}to}, \texttt{choose\_\allowbreak{}dialogue\_\allowbreak{}option}, \texttt{end\_\allowbreak{}dialogue} & Speaker, line, indexed options, quest/shop changes \\
Collection and interaction & \texttt{fetch\_\allowbreak{}item}, \texttt{loot\_\allowbreak{}all}, \texttt{loot\_\allowbreak{}item}, \texttt{interact\_\allowbreak{}with} & Actual transfers, remaining loot, object/dialogue results \\
Commerce and equipment & \texttt{buy\_\allowbreak{}item}, \texttt{sell\_\allowbreak{}item}, \texttt{equip\_\allowbreak{}item} & Actual quantity, coin balance, stock, equipped state \\
Combat and progression & \texttt{attack\_\allowbreak{}nearest}, \texttt{get\_\allowbreak{}combat\_\allowbreak{}state}, \texttt{combat\_\allowbreak{}action}, \texttt{get\_\allowbreak{}level\_\allowbreak{}up\_\allowbreak{}choices}, \texttt{choose\_\allowbreak{}level\_\allowbreak{}up\_\allowbreak{}ability} & Battle phase/outcome, legal actions, events/deltas, offered upgrades \\
Visual inspection & \texttt{look}, \texttt{inspect\_\allowbreak{}map}, \texttt{read\_\allowbreak{}image} & Fresh views, authored map images, saved native vision inputs \\
Composition and continuation & \texttt{run\_\allowbreak{}code}, \texttt{set\_\allowbreak{}goal}, \texttt{complete\_\allowbreak{}goal} & Script findings/errors and active-objective state \\
Bug reporting & \texttt{note\_\allowbreak{}issue} & Saved expected/observed behavior with engine-derived evidence \\
\bottomrule
\end{tabularx}
\end{table*}
\subsubsection{Execution Feedback}
\label{sec:org083a650}

Critical to the tool design for our agent is feedback from executed tools, which are used to inform the agent about progress it makes inside the environment. For example, if the agent attempts to move to a location, but is physically blocked by a wall, then the issuing of a command to navigate to \texttt{(x,y,z)} is insufficient proof that the agent has arrived at \texttt{(x,y,z)}. The agent will, in this case, need to make another attempt at navigating to the target location. Each tool accessible to the agent will produce feedback which can be fed back into the agent, necessary for both perception and action tools.

The two components which make this possible in the Dogwood environment are the Python runtime and the Godot executor. The Python runtime will queue commands for the Godot executor, and will wait for a response from the executor before proceeding. When the action completes or fails, the executor reports back the unique action ID, the outcome, and an updated world snapshot. The Python runtime then consumes this world state, releases the waiting tool call, and produces a structured output which can be consumed by the agent for its next action. This allows the agent to see if the tool timed out, completed, returned any useful information (e.g. observed dialogue or nearby NPCs), and so on.
\subsubsection{Structured Composition with Generated Code}
\label{sec:org3e27357}

Mindcraft's \texttt{!newAction} sends a plan to a JavaScript coding loop with retrieved skill documentation. Following CodeAct~\cite{codeact}, we adapt this idea in the Dogwood environment as programmatic tool calling: \texttt{run\_code} executes Python written by the game-playing agent, composing the available tools into larger blocks of code rather than executing them command-by-command. Because tools emit dictionaries (containing completion feedback for executed actions, or observations for information-gathering tools), these dictionaries can be used as gates or inputs for future tools. For example, a code block could terminate when a tool call fails, or the agent could write a loop to talk to all NPCs returned by a \texttt{get\_nearby\_entities} tool. 

To illustrate a larger example, consider the case where the agent needs to locate a specific type of enemy. The agent can open the map, and generate a list of candidate locations where this enemy might be located. Then the agent could compose a block of code which travels to each of these locations, searches the area for all entities, and returns with the current location if the target enemy is located during the execution of this block. This allows the agent to perform more human-like exploration inside a large environment via composed tools, rather than equipping the agent with an unrealistic search radius or entity location lookups.

We note that the design of the Dogwood environment places greater emphasis upon this exploration component compared to Voyager, which relies on the Mineflayer \texttt{findBlocks} tool. ~\cite{voyager} This allows it to peer through obstructed ground to notice valuable targets within a 3D radius which would not be visible to a human player, for example seeing iron ore 20 blocks underneath the current position. With a comparatively tight search radius, the agent must explore the map to locate important resources via tool composition, rather than relying upon an oracle. 
\subsubsection{Persistent Objectives and Bug Reporting}
\label{sec:org7123d5d}

The \texttt{set\_goal} tool records an objective that persists across agent turns. When a turn ends with the objective still active, the runtime prompts the agent again, asking it to either call \texttt{complete\_goal} to set the objective to complete, or to resume making progress on the goal. This simple loop allows for long-horizon, multi-stage objectives to be persistently pursued without further human prompting beyond the initial natural language goal (e.g. "win the game"). Judging goal completion can be checked against emitted game state when applicable (e.g. "quest is marked completed").

Bug finding uses the same tools as general gameplay, and adds \texttt{note\_issue(expected, observed, blocked)}. This tool emits a ticket which saves the context surrounding an issue observed in gameplay, along with optional captured visual evidence like screenshots. If the issue is logged as \texttt{blocked}, e.g. in a case where the game has crashed or play cannot continue, the session is subsequently closed.
\subsubsection{Gameplay Persona Conditioning}
\label{sec:org90d3a9f}

A key feature of our approach is the ability to generate synthetic gameplay trajectories which are described in natural language. This is straightforward to understand for concrete objectives that need to be completed (e.g. "complete the hero's path questline") but critically also can describe \emph{how} we accomplish that goal (e.g. "grind as much experience as possible through combat").

Our system, via Hermes Agent, can accept a configurable \texttt{SOUL.md} file, which can define the agent's core personality or behaviors. For open-ended exploration tasks, this allows us to condition the agent's decisions on imagined player behaviors or archetypes, in order to more broadly cover the search space than would be explored by default. For example, a \texttt{SOUL.md} file could instruct the agent to attack lots of NPCs, to try to avoid combat altogether, to focus on gathering resources or trading for lots of gold, and so on. Similar results can be observed by including these instructions in a system prompt, or elsewhere in the assembled prompt.

\begin{figure*}[t]
\centering
\includegraphics[width=\textwidth]{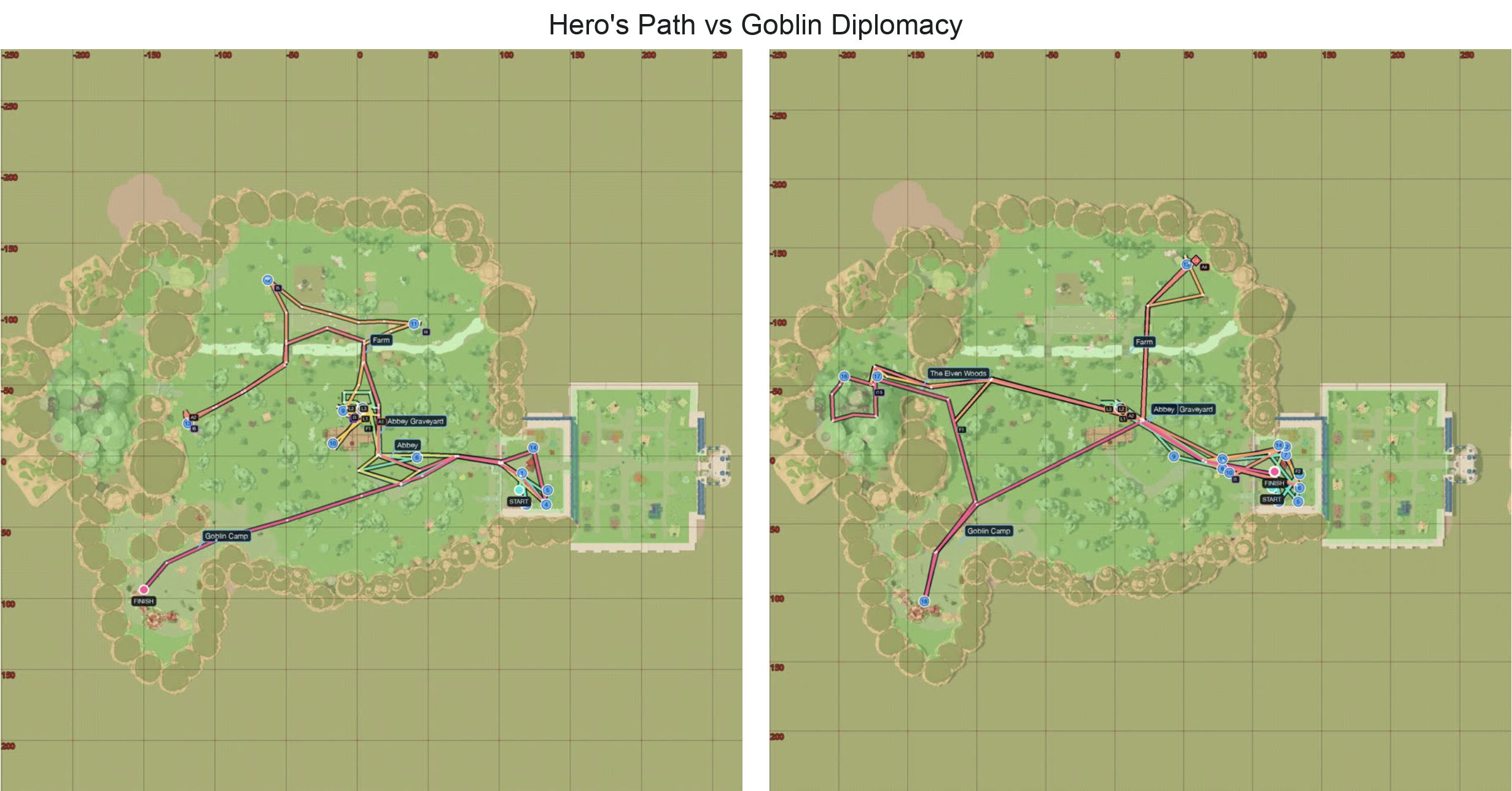}
\caption{\label{fig:orgc4c7b8c}Hero's Path vs Goblin Diplomacy. Left: the agent is instructed to complete the game through heroic conquest, leading to a dramatic confrontation with the Goblin King. Right: the agent is instructed to beat the game while avoiding unneeded conflict, and ends the game through diplomacy.}
\end{figure*}

We find that we are able to produce very different trajectories through these simple prompting changes, which further adds an additional modular element to our system. Figure \ref{fig:orgc4c7b8c} illustrates the different routes and exploration orders in two completed persona conditioning trials, yielding different quest lines completed en route to completing the game. 

While we do not claim that this approach can faithfully emulate human behavior in games, the ability to demonstrate these differentiated synthetic players may enable developers to identify areas of their game which are unintentionally thin, portions of the audience who may be underserved, or gaps which appear for one playstyle and not another. In this way, Buddy could contribute in the playtesting process in some basic capacity even for more subjective questions.  
\subsection{Multimodal / Keyboard-and-Mouse Capabilities}
\label{sec:orga014ccd}

An important capability of many modern LLMs is multimodal understanding. While privileged game state is highly useful, there are many times where it is difficult to infer what is actually happening on screen just from gamestate. A highly useful capability, therefore, is the ability to take a snapshot of the game and to provide this to the language model agent.

We do not consider Buddy to be a "from pixels" approach to game-solving, since the primary mechanism for perceiving the world is via game state. This is for practical purposes: Buddy is intended to be useful for game development, and game developers can provide gamestate to an agent without too much difficulty. For our intended application, we do not consider a pure from-pixel approach to be necessary to demonstrate value; however, a pure no-pixel approach can be similarly impractical. We find that the agent will typically understand when it needs to check what is happening visually, and will leverage a "take screenshot" tool in moderation when it gets stuck. 

Likewise, popular agent harnesses like Hermes Agent ship with computer use capabilities, which could allow our agent to pilot the game with keyboard and mouse actions as if it were a human player.~\cite{hermescomputeruse} For our Dogwood runs, this is a feature which we ablate by default. We do not consider Buddy to be a "keyboard-and-mouse emulation" approach to game-solving, since as with multimodal perception, we primarily interact with the game through action commands. However, as with multimodal perception, we see no need to exclude this capability from the framework altogether, since it ships natively with a popular agent harness, and the intent is to be able to use these harnesses in a modular way. 

\begin{table*}[!t]
\centering
\footnotesize
\setlength{\tabcolsep}{3pt}
\caption{Our results with different models, pooled across harnesses. \emph{Completed} counts full solves over the attempts listed in Appendix A; \emph{Quest progress} is the best verified result across those attempts. Total tokens report median usage across those attempts; input and output medians use only attempts with available token breakdowns.}
\label{tab:token-usage}
\begin{tabularx}{\textwidth}{@{}l c c *{4}{>{\raggedleft\arraybackslash}X}@{}}
\toprule
Model & Completed & \shortstack{Quest\\progress} & Input tokens & Output tokens & Total tokens & Cost (USD) \\
\midrule
\texttt{glm-5.3-flash}               & 1/2 & 18/18 & 11.63M & 14.11K & 21.72M & 0.490 \\
\texttt{gpt-5.6-luna}                & 3/3 & 18/18 & 22.70M & 32.20K & 29.61M & 0.690 \\
\texttt{gpt-6-astra}                 & 2/2 & 18/18 & 8.126M & 11.85K & 10.57M & 13.601 \\
\texttt{gemini-3.8-flash}            & 4/4 & 18/18 & 5.418M & 10.56K & 5.860M & 0.934 \\
\texttt{deepseek-v4.1-flash}         & 2/2 & 18/18 & 16.56M & 43.77K & 36.60M & 0.538 \\
\bottomrule
\end{tabularx}
\par\smallskip
\begin{minipage}{\textwidth}
\footnotesize
Input tokens include cached prompt tokens.
\end{minipage}
\par\medskip
\caption{Harness results. \emph{Completed} counts full solves over the attempts listed in Appendix A; \emph{Quest progress} is the best verified result across those attempts. Token columns report median usage for each model--harness pair. Providers and reasoning settings vary across runs, as listed in Appendix A.}
\label{tab:harness}
\begin{tabularx}{\textwidth}{@{}l *{6}{>{\raggedleft\arraybackslash}X}@{}}
\toprule
 & \multicolumn{3}{c}{Hermes} & \multicolumn{3}{c}{Codex} \\
\cmidrule(lr){2-4} \cmidrule(lr){5-7}
Model & Completed & \shortstack[r]{Quest\\progress} & Total tokens & Completed & \shortstack[r]{Quest\\progress} & Total tokens \\
\midrule
\texttt{glm-5.3-flash}               & 0/1 & 13/18 & 11.64M & 1/1 & 18/18 & 31.8M \\
\texttt{gpt-5.6-luna}                & 2/2 & 18/18 & 22.73M & 1/1 & 18/18 & 43.2M \\
\texttt{gpt-6-astra}                 & 1/1 & 18/18 & 8.137M & 1/1 & 18/18 & 13.0M \\
\texttt{gemini-3.8-flash}            & 3/3 & 18/18 & 5.427M & 1/1 & 18/18 & 10.5M \\
\texttt{deepseek-v4.1-flash}         & 1/1 & 18/18 & 16.60M & 1/1 & 18/18 & 56.6M \\
\bottomrule
\end{tabularx}
\end{table*}
\subsection{Trace-to-HTN Conversion in Dogwood}
\label{sec:orgba23e9c}

After Buddy completes an objective, a separate coding agent session converts the successful trace into a hierarchical task network (HTN)~\cite{htn}. The coding agent also receives quest dependencies and world context pulled from the game source, including default NPC and resource locations. The resulting plan removes exploration, information gathering, non-repeatable actions, map and screenshot usage, and so on.

We use the Godot Fluid HTN addon as the GDScript execution layer in Godot 4.6.~\cite{godotfluidhtn} An LLM is used to produce a task tree in JSON comprised of built-in game actions and world-state checks, which is then interpreted by the GDScript runtime. 

We use GPT-6 Astra as the conversion model. The agent generates a candidate tree, executes it, inspects failures, and revises missing or inapplicable methods until the tree successfully completes the original objective. Once the objective is completed, the resulting HTN can complete the task without further language model calls. Unlike approaches that learn a general PDDL action model from game logs before planning test scripts, this procedure constructs a task-specific artifact from a successful execution trace and engine context.~\cite{planning} Appendix \ref{sec:htn-execution} describes the representation and execution procedure.
\section{Results}
\label{sec:org5900e52}

\subsection{Autonomous Game Solving}
\label{sec:org60b6d09}

Our Buddy is capable of completing Dogwood's entire main quest line autonomously with no prior knowledge or instruction beyond the initial goal of "win the game." It will play unassisted for 45+ minutes to accomplish this, completing multiple quests in sequence across the environment, and exploring the map to unseen locations in order to progress. All of this can be done with smaller models like gpt-5.6-luna, with a median per-run cost of \$0.69. The game-solving benchmark runs skip the combat encounters for brevity, since these are trivially solved by grinding if encounters are too difficult. A case study on completing the game with combat enabled can be found in Appendix \ref{sec:combat-case-study}.
\subsubsection{Model Flexibility}
\label{sec:orgecbd129}

An important component of our work is that it's flexible to future improvements in language model technology across the Pareto frontier. As language models continue to improve, we can simply swap out the model that is used to power the system. Table \ref{tab:token-usage} reports token usage and cost for several game-solving attempts across a range of off-the-shelf models.
\subsubsection{Harness Flexibility}
\label{sec:org853ae69}

Our system was primarily engineered around the Hermes Agent harness, but this is another component which can be swapped out for another harness without substantial difficulty. As a demonstration of this, we also implement a setting which allows the harness to be swapped with Codex CLI while retaining all the same tools and abilities. The Codex-backed agent is also capable of completing the entire game, composing tool calls, switching between models, and recovering from errors.

Below we show some straightforward results using the same models with Hermes and Codex backends. Table \ref{tab:harness} compares completion rate and total token usage across the two harnesses.
\subsection{Bug Finding}
\label{sec:orgdd2c334}

Buddy is capable of logging difficulties it encountered in playtesting. It does this by invoking a tool which allows it to populate a 3-tuple: a string describing the expected behavior, a string describing the observed behavior, and a bool describing whether or not the difficulty was a blocking bug or not. This way, we can create detailed reports after playthroughs of the anomalous behaviors we encountered, and can determine if they need addressing. 

\begin{figure}[t]
\centering
\includegraphics[width=\columnwidth]{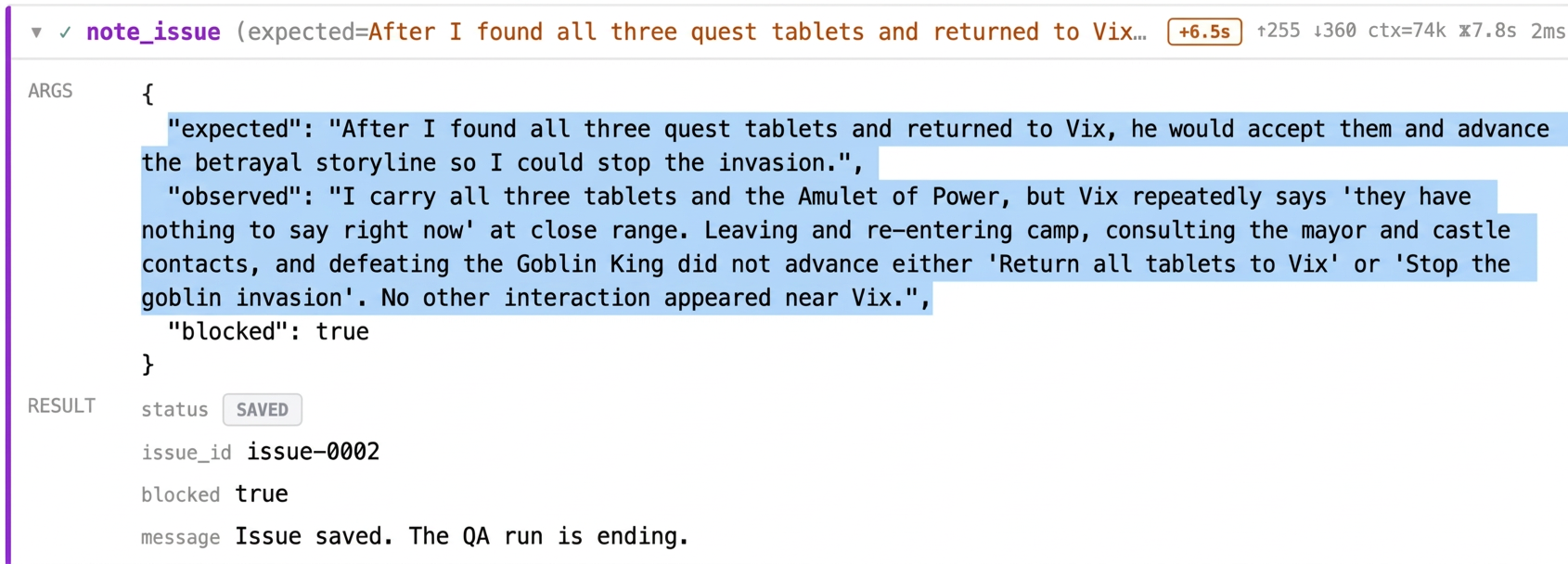}
\caption{\label{fig:org1528993}An example of Buddy logging a bug ticket.}
\end{figure}

Many of the difficulties encountered by our Buddy in the typical Dogwood playthrough involve getting stuck somewhere due to broken navmesh grids. These are a clear example of a bug that should be repaired, since NPC characters navigating the same area can encounter the same navigation failure. Broken navmesh grids are an example of a \emph{recoverable bug} since frequently the agent can find some path from point A to point B and still complete the task. The agent can label this location as a bug, file a ticket, and then continue trying to solve the task through other means. One example Codex run got stuck between a static cart and a large rock due to the collision of the rock and the navmesh interacting unfavorably. After trying several different methods, the agent was able to free itself and resume the quest. This nicely illustrates a desirable behavior in autonomous playtesting: an unintended issue can be solved through thrashing or clever rerouting, and then the original objective can still be pursued.

\begin{figure}[t]
\centering
\includegraphics[width=\columnwidth]{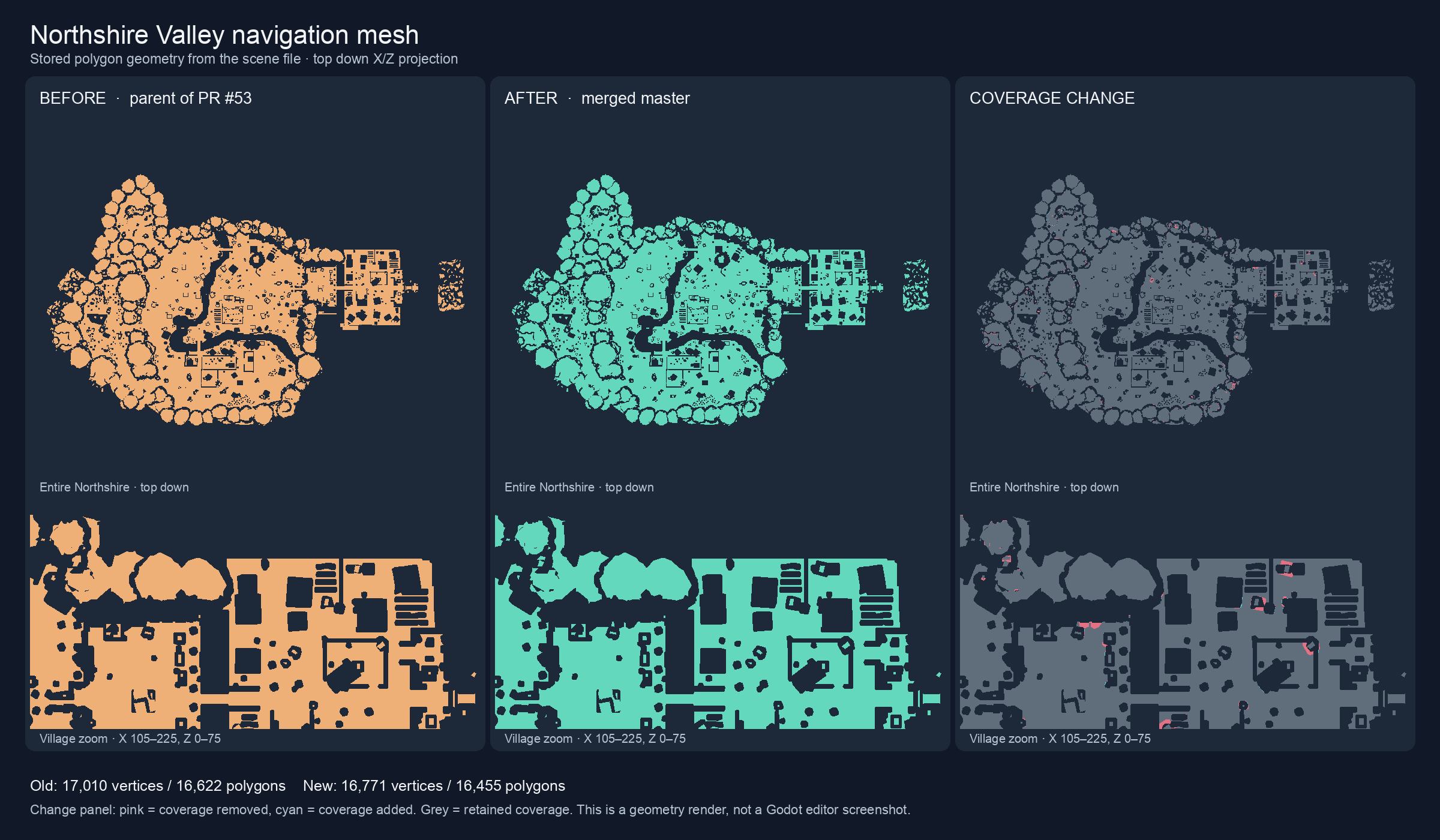}
\caption{\label{fig:org9729e7e}Buddy's difficulty with navmesh islands was one example of a usually-recoverable bug which was brought to our attention through automated playtesting, which was then repaired.}
\end{figure}

\begin{figure}[t]
\centering
\input{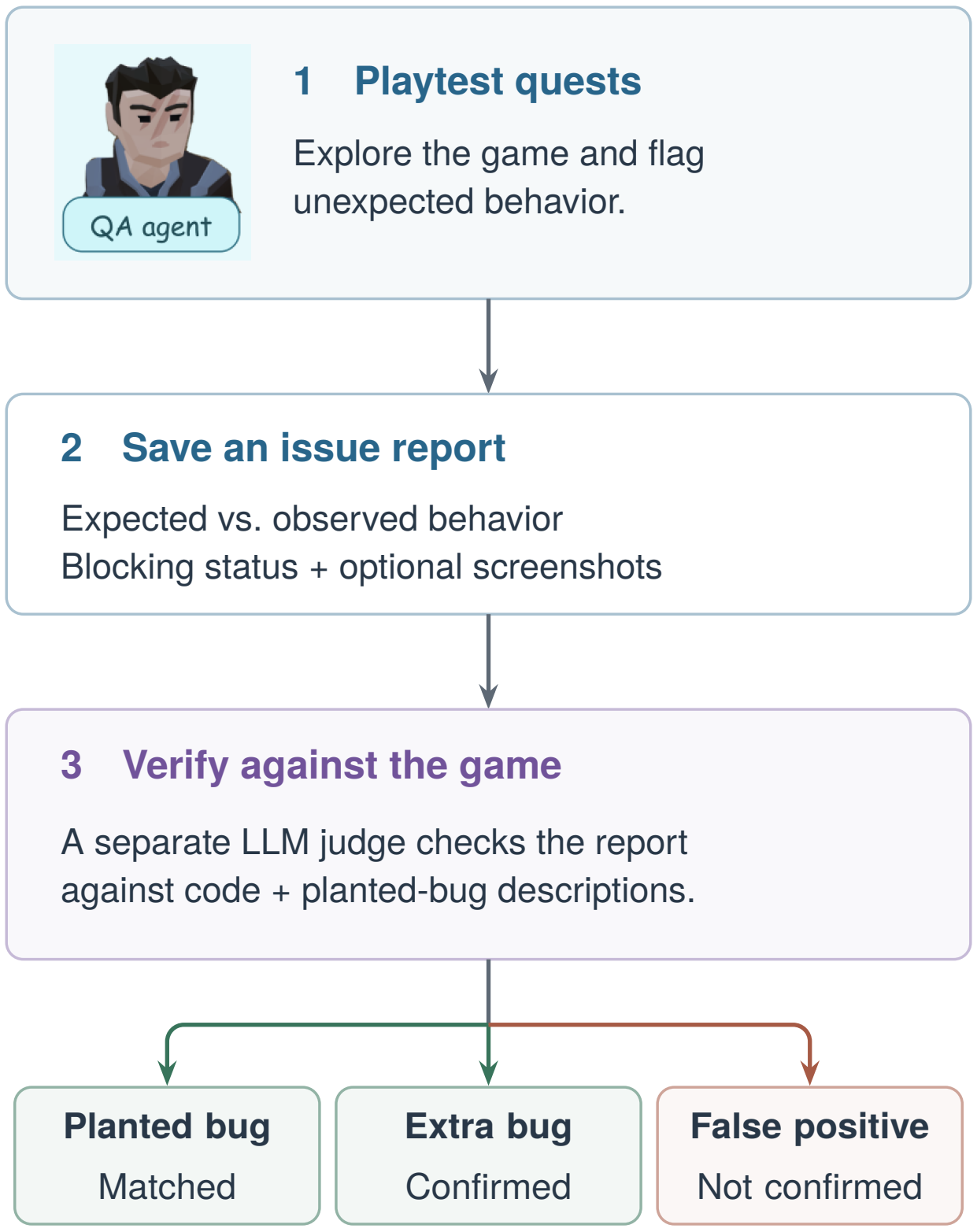}
\caption{Autonomous bug hunting: a QA agent reports unexpected behavior, and a separate judge verifies each report using game code and planted-bug descriptions.}
\label{fig:bug_logging}
\end{figure}

Quests which are not completable are an example of a \emph{blocking bug}. When encountering a situation like this, Buddy will first make some attempts to route around the apparent issue (e.g. by leaving and returning, by trying a few more times, by attempting to reload the zone, etc). A blocking bug is typically more severe than a recoverable one, since the playthrough will not be able to proceed past it. 
\subsubsection{BugGym}
\label{sec:org7ba59d6}

Following TITAN~\cite{titan}, we construct BugGym, a proxy environment for Dogwood containing 48 quest families, each with one clean and one planted-bug variant, yielding 96 trials per model.

\textbf{Bug Taxonomy}

To ensure the types of bugs in the BugGym cover a wide range of possible issues, we categorize bugs in the environment into 7 broad types:

\begin{enumerate}
\item Cross-Quest Interference - quests which cause issues if overlapped.
\item Irreversible Player Actions - permitted actions which can crash or softlock the game, like destroying a quest item.
\item Entity Placement - key objectives not accessible or instantiated in the world.
\item Wrong Directions - quest instructions do not match the actual task
\item Balance Concerns - quests which can be completed through unintentionally tedious means, e.g. a key item drop rate of 0.5\% instead of 50\%.
\item Visual Immersion Breaks - NPCs being the wrong type, objects which are misplaced or floating, etc.
\item Too Much Information - hidden information presented too early, locked objectives listed before they unlock, quest log which spoils a twist, etc.
\end{enumerate}

We construct the BugGym to have examples along each of these categories. 

\textbf{Evaluation}

Models are assigned a quest or related quest bundle and instructed to complete it while reporting directly observed problems through the \texttt{note\_issue} tool. The gameplay agent has no access to game source, planted-bug descriptions, or other runs.

To verify each issue raised by the agent, we use LLM-as-judge~\cite{llmjudge} using gpt-6.1 Sol high as the adjudicator, with full read access to the planted-bug answer key, the game source code, and the full trace emitted by the playtesting agent. Issues which match the description of a planted bug are marked as true positives. For issues which do not match the description of a planted bug, following Bug Hunt Bench~\cite{bughuntbench} we direct the judge to determine if this bug exists as an unintended bug in our environment. If the bug is confirmed, we credit the agent with an "extra" bug, which we report as an additional metric in Table \ref{tab:buggym}. Reports contradicted by the evidence are marked as false positives, whereas unresolved reports are marked inconclusive.

ReviewBench similarly credits findings outside its reference set in augmented precision.~\cite{reviewbench} CyberGym-E2E distinguishes finding and patching an alternative vulnerability from finding the intended ground-truth vulnerability.~\cite{cybergyme2e}

\textbf{Metrics.}

Let \(N\) be the number of in-scope issue reports after deduplication within each trial, \(N_s\) the number accepted as reports of planted defects, \(N_e\) the number accepted as reports of unintended defects, \(N_f\) the number classified as false positives, and \(N_u\) the number remaining inconclusive, so that \(N=N_s+N_e+N_f+N_u\). We report:

\begin{align*}
P_{\mathrm{aug}} &= \frac{N_s+N_e}{N}, &
\mathrm{FDR} &= \frac{N_f}{N}, \\
R_{\mathrm{seed}} &= \frac{D_s}{B}, &
\mathrm{Extras} &= D_e,
\end{align*}

where \(B\) is the number of distinct planted defects in the evaluation manifest, and \(D_s\) and \(D_e\) count distinct planted and unintended defects identified, respectively. Repeated reports of the same defect do not increase recall or Extras. Precision is report-level and includes accepted unintended findings; Extras does not increase planted-bug recall. If no reports are submitted, precision and FDR are undefined and are reported as ``N/A,'' not zero.

FDR is the fraction of reports classified as false positives, equal to \(1-P_{\mathrm{aug}}-N_u/N\). It is not the false-positive rate \(FP/(FP+TN)\), which would require a defined population of negative decisions. A quest without a deliberately planted bug does not by itself supply that population. We retain the fixed planted-defect denominator for recall because the complete set of unintended defects is unknown.

More details about the evaluation and setup can be found in Appendix \ref{sec:buggym-protocol}.

\textbf{Results}

Table \ref{tab:buggym} reports the BugGym results. Astra and Argon each find 31 of 48 planted defects (64.6\% recall), while Astra has the highest augmented precision (98.5\%) and finds 11 distinct unintended bugs.

\begin{table*}[t]
\centering
\footnotesize
\setlength{\tabcolsep}{4pt}
\caption{BugGym results pooled across 96 trials per model. Recall counts distinct planted defects; precision includes judge-accepted planted and unintended findings. Extras counts distinct unintended defects. (Appendix~\ref{sec:buggym-protocol}).}
\label{tab:buggym}
\begin{tabularx}{\textwidth}{@{}>{\raggedright\arraybackslash}p{2.7cm} *{6}{>{\raggedleft\arraybackslash}X}@{}}
\toprule
Model & Trials & Reports & \shortstack[r]{Seeded\\recall (\%)} & \shortstack[r]{Augmented\\precision (\%)} & Extras & \shortstack[r]{FDR\\(\%)} \\
\midrule
GPT-6 Luna & 96 & 36 & 43.8 & 91.7 & 3 & 8.3 \\
GPT-6.1 Sol & 96 & 74 & 54.2 & 95.9 & 10 & 2.7 \\
GPT-6 Astra & 96 & 67 & 64.6 & 98.5 & 11 & 0.0 \\
Gemini 4 Argon & 96 & 56 & 64.6 & 89.3 & 10 & 10.7 \\
\bottomrule
\end{tabularx}
\par\smallskip
\begin{minipage}{\textwidth}
\footnotesize
Sol and Astra each have one inconclusive report, retained in the denominator but counted as neither a confirmed bug nor a false positive.
\end{minipage}
\end{table*}
\subsection{Trace Conversion}
\label{sec:org9021a48}

\begin{figure}[t]
\centering
\includegraphics[width=\columnwidth]{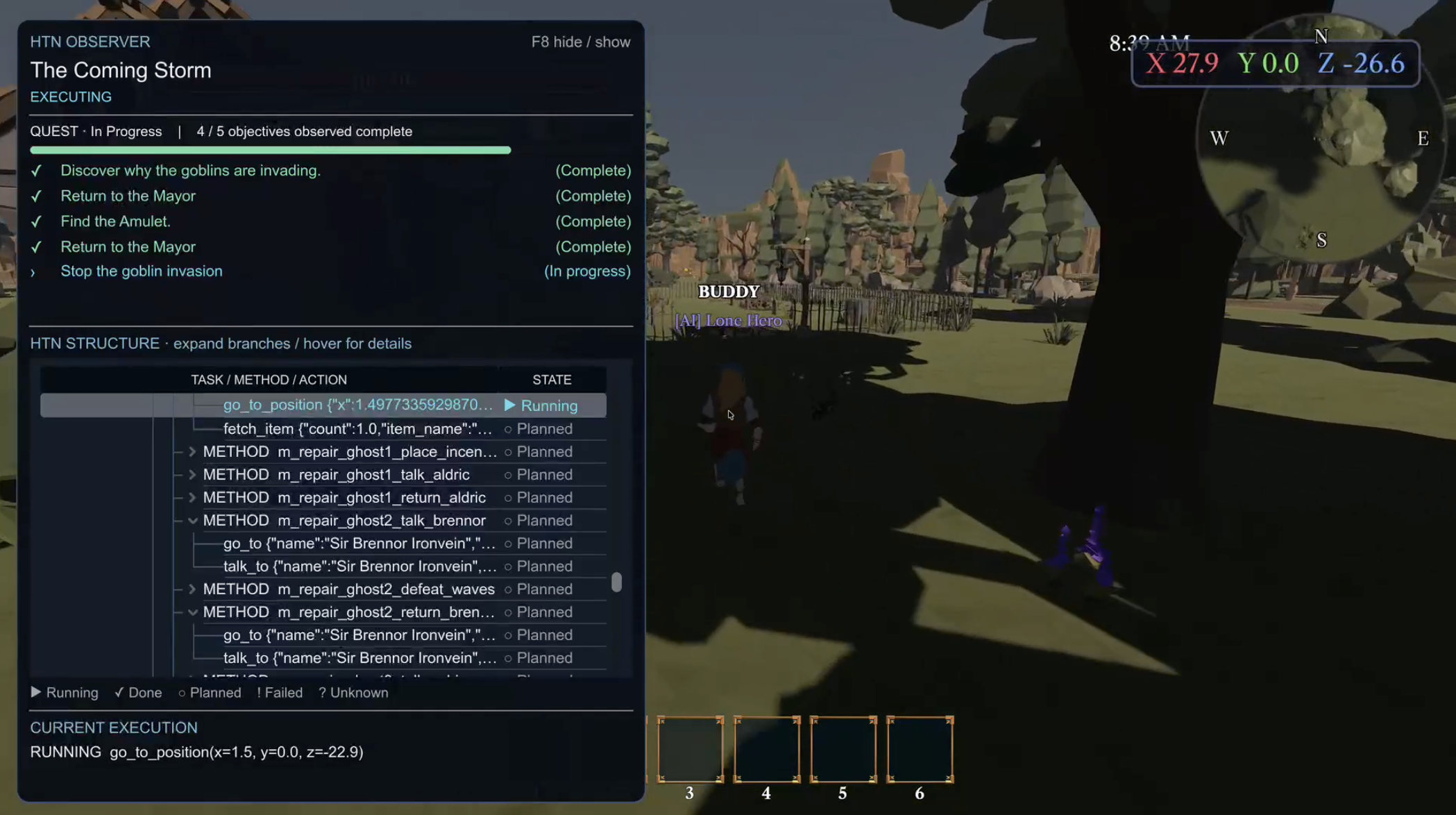}
\caption{\label{fig:org4693ee0}A converted HTN from a full game-solving trace, completing the game with no language model invocation.}
\end{figure}

We converted successful gameplay traces into reusable JSON HTNs that complete the objective without language-model calls during replay. Figure \ref{fig:org4693ee0} illustrates execution of one such artifact, a full-game end-to-end gameplay trace which was converted into an HTN that solves the same objective in 7.6 minutes wall-clock time. Observed failures in this process included reliance on recorded coordinates for moving NPCs and references to particular inventory slots or instance identifiers. These examples motivate the execution-and-repair loop described in the architecture section.

\section{Discussion and Conclusion}
\label{sec:orgf8255c2}

Improvements in LLM capability and cost can make autonomous game playtesting more practical. Increased performance in visual perception, computer use, code understanding, long context, error reporting, and multi-turn can all contribute to LLM-driven autonomous playtesting for game development continuing to grow in viability. 

At the core, our claims are very simple:

\begin{enumerate}
\item Off-the-shelf models are capable of playing unreleased games, even without supplied walkthroughs.
\item Popular agent harnesses can be repurposed for game-playing agents with minimal modification.
\item These systems are useful and affordable for game playtesting.
\item This can all be straightforwardly assembled by game studios without training or research teams.
\end{enumerate}

We also believe this approach is relatively future-proof and allows game developers to reap the benefits of the continued pace of improvement in language models without locking them behind any ecosystem tied to a particular model. If a new model comes out which is more capable, you can swap it in to make your playtesting loop more capable. If a new model comes out which is cheaper, you can swap it in to make your playtesting loop cheaper. If a new harness comes out which enables faster and more robust computer use, you can swap it in to make your system better.

We point out that many playtesting tasks lie outside the domain of this particular system, especially when using game-state tools instead of pure computer use. While we can test many gameplay elements this way, a tradeoff exists between testing every player-facing input and UI path, and simplifying some interaction primitives for cost and speed (e.g. "view dialogue tree, select option 1" via gamestate vs "view in-game screenshot, send click event at location to accept option 1").

Finally, we note that battles are skipped in the benchmark runs for brevity since difficult encounters are easily solvable in our environment by grinding experience until they can be completed. A similar strategy appears in Claude Fable 5's completion of Pokemon FireRed, where it levels its starter Pokemon to level 79 and uses the other Pokemon just to buy turns to revive the starter.~\cite{claudefable5} A systematic evaluation of combat strategy and efficiency is left to future work, but a case study with combat enabled can be found in Appendix \ref{sec:combat-case-study}. Our evaluation is performed on a relatively small, held-out environment with relatively low sample sizes across model-harness pairs, as the intent of our results was to demonstrate feasibility rather than benchmark the efficiency of models or harnesses.

Our hope is that our system hits a practical sweet spot: easy to assemble, easy to improve, inexpensive to run and maintain, and targeting the most mundane and routine playtesting tasks. With these tasks automated, human in-game time can be more pointedly focused on tastemaking and actually improving the game, rather than just making sure everything is connected together.
\section{Acknowledgements}
\label{sec:org1044bac}

We thank George Yang for his work on implementing the initial HTN solver in the Dogwood environment. We thank Timbre Games for their work developing the Dogwood environment. We thank Wesley Kerr for mentorship and guidance over this project.

\clearpage
\appendix
\counterwithin*{table}{section}
\renewcommand{\thetable}{\thesection\arabic{table}}
\section{Benchmark Tables}
\label{sec:orgc9954ca}

All attempts with all runs use the same initial prompting: a provided goal string of "win the game" with no walkthrough or additional information. A full solve involves passing 18 quest milestones: 11 in the main quest, and an additional 7 in the Hero's Path quest. These runs include a fully invincible player who instantly wins combat.

\begin{table*}[t]
\centering
\footnotesize
\setlength{\tabcolsep}{4pt}
\caption{Per-run outcomes and execution settings. R1--R8 and R13 use Hermes; the remaining runs use Codex. Quest progress lists the number of completed milestones. Provider identifies the service used to access the model; effort is the requested reasoning setting, with xhigh denoting extra-high effort. Incomplete-run wall times end at the recorded stop.}
\label{tab:run-outcomes}
\begin{tabularx}{\textwidth}{@{}l >{\raggedright\arraybackslash}X l l >{\raggedright\arraybackslash}p{2.7cm} c r r@{}}
\toprule
Run & Model & Provider & Effort & Outcome & \shortstack{Quest\\progress} & \shortstack[r]{Time\\(min)} & \shortstack[r]{Tool\\calls} \\
\midrule
R1 & Gemini 3.8 Flash & OpenRouter & medium & Complete & \shortstack{18/18\\(11/11 + 7/7)} & 21.9 & 201 \\
R2 & Gemini 3.8 Flash & OpenRouter & medium & Complete & \shortstack{18/18\\(11/11 + 7/7)} & 28.7 & 287 \\
R3 & Gemini 3.8 Flash & OpenRouter & medium & Complete & \shortstack{18/18\\(11/11 + 7/7)} & 28.0 & 208 \\
R4 & GPT-5.6 Luna & OpenAI & xhigh & Complete & \shortstack{18/18\\(11/11 + 7/7)} & 38.4 & 195 \\
R5 & GPT-5.6 Luna & OpenAI & xhigh & Complete & \shortstack{18/18\\(11/11 + 7/7)} & 71.7 & 276 \\
R6 & GLM 5.3 Flash & OpenRouter & medium & Stopped on Hero's Path & \shortstack{13/18\\(10/11 + 3/7)} & 79.6 & 215 \\
R7 & DeepSeek V4 Flash Vision & OpenRouter & medium & Stopped on Hero's Path & \shortstack{14/18\\(10/11 + 4/7)} & 72.7 & 376 \\
R8 & GPT-6 Astra & OpenAI & medium & Complete & \shortstack{18/18\\(11/11 + 7/7)} & 28.4 & 240 \\
R9 & Gemini 3.8 Flash & OpenRouter & xhigh & Complete (Codex) & \shortstack{18/18\\(11/11 + 7/7)} & 30.6 & 234 \\
R10 & GPT-5.6 Luna & OpenRouter & xhigh & Complete (Codex) & \shortstack{18/18\\(11/11 + 7/7)} & 116.1 & 697 \\
R11 & GLM 5.3 Flash & OpenRouter & xhigh & Complete (Codex) & \shortstack{18/18\\(11/11 + 7/7)} & 155.45 & 971 \\
R12 & DeepSeek V4.1 Flash & OpenRouter & xhigh & Complete (Codex) & \shortstack{18/18\\(11/11 + 7/7)} & 98.8 & 1047 \\
R13 & DeepSeek V4.1 Flash & OpenRouter & xhigh & Complete (Hermes) & \shortstack{18/18\\(11/11 + 7/7)} & 36.8 & 335 \\
R14 & GPT-6 Astra & OpenRouter & medium & Complete (Codex) & \shortstack{18/18\\(11/11 + 7/7)} & 24.4 & 346 \\
\bottomrule
\end{tabularx}
\end{table*}
\section{BugGym Evaluation Protocol}
\label{sec:org7a61795}
\label{sec:buggym-protocol}
\subsection{Scenario Design}
\label{sec:orgd9e613e}

BugGym contains 48 quest families, each with one clean and one planted-bug variant sharing the same scenario. Cross-quest scenarios include the necessary bundle of interacting quests. Each model is evaluated on 96 valid scenario-variant trials, with 48 planted bugs available. Clean variants have no deliberately planted defects, but may still contain unintended issues, which we report as "extra" when encountered by gameplay agents. 

The planted set contains 37 "functional bugs" and 11 "design concerns". Functional bugs break intended mechanics, progression, or state. Design concerns involve communication, balance, or presentation, such as misleading directions or premature spoilers. Both contribute to the reported metrics, as do confirmed unintended findings of either kind.
\subsection{Gameplay Protocol}
\label{sec:orgf89a2a5}

The runner uses Godot 4.6.2 and the Hermes Agent harness. Each trial starts a fresh game and agent session with isolated saves, a cleared scratchpad, and scenario-specific quest resources and allowed scenes. The player receives an assignment naming the quest giver and starting location, and no additional context or access to game source. It pursues the assignment, attempts to recover from errors, and reports observed problems through \texttt{note\_issue}. Luna and Sol use xhigh reasoning with 30-minute gameplay limits; Astra and Argon use high reasoning with 45-minute limits.

Completion requires every assigned quest to reach its engine-recorded completion or turn-in state, rather than the agent merely calling \texttt{complete\_goal}. Trials also end when an issue is marked as "blocking" or when the agent passes the gameplay time limit. A timeout does not erase confirmed findings. Invalid launches, missing traces, and unverified model or source identities are excluded from scoring rather than treated as bug-free trials.
\subsection{Access and Evidence}
\label{sec:orgf3e21ea}

The gameplay agent's file and code tools cannot access game source, quest resources, answer keys, or other runs. Permitted access covers session scratch space, game screenshots, and required runtime libraries. Generated Python is confined to approved files and gameplay-tool calls.

Each \texttt{note\_issue} record combines the agent's expected/observed descriptions and blocking flag with engine-derived position and quest state, a timestamp, and the latest available screenshot. After gameplay exits, the pipeline seals reports, screenshots, logs, traces, and adjudication inputs, recording file hashes and source fingerprints. The adjudicator inspects saved evidence and game/harness code read-only; it does not replay gameplay or repair the game.
\subsection{Adjudication and Scoring}
\label{sec:org7283069}

The judge evaluates submitted in-game claims, not additional defects it discovers while inspecting code. Matching a planted-bug description requires evidence of the reported problem, not merely the defect's presence in the answer key. Tool-only failures are out of scope and excluded from the report denominator. Confirmed reported defects outside the planted set receive credit, including those found in clean variants.

Metrics pool the final reviewed findings after deduplication within each trial. Repeated unintended findings across trials contribute to report-level precision, while Extras counts distinct underlying causes. Planted-bug recall uses all 48 planted defects, including those not encountered during valid gameplay. Reports left inconclusive after re-review remain in the precision denominator without positive credit and are not counted as false positives. 
\section{HTN Implementation}
\label{sec:orgef45698}
\label{sec:htn-execution}
\subsection{Representation}
\label{sec:org8ce573a}

The HTN plans use JSON schema version 2 and execute through Godot Fluid HTN v0.4.1. A plan identifies a \texttt{root\_task} and a collection of tasks, each with a \texttt{complete\_when} condition and alternative methods. Methods contain preconditions and subtasks, ultimately invoking 22 built-in game actions and 32 world-state checks. Validated plans are saved for reuse on the same quest, rather than as parameterized skills shared across quests.
\subsection{Context}
\label{sec:orge64bf6c}

Alongside the successful gameplay trace, the model receives the quest and sub-quest dependency graph, unlock requirements, NPC locations and offered quests, map regions, item sources and loot tables, and so on. Live observations include nearby entities and interactables within 50m, inventory, dialogue options, shop contents, and recent action results.
\subsection{Repair Loop}
\label{sec:org3de142a}

After each action, the runtime reads actual game state and traverses from the root to the first unfinished task, selecting a method whose preconditions hold. It does not rely on forward-simulated action effects. Checks distinguish \texttt{TRUE}, \texttt{FALSE}, and \texttt{UNKNOWN}, so unobserved or still-spawning entities are not treated as evidence of completion. Target coordinates and dialogue choices are checked against observed game state before execution. During construction, an empty method list or a lack of applicable methods triggers a model call with the current tree, blocked task, and latest observations. The model patches the blocked task and execution resumes. This iteration produces the saved plan used for model-free replay.
\section{Combat-Enabled Case Study}
\label{sec:org1b9c16a}
\label{sec:combat-case-study}
\subsection{Setup and Outcome}
\label{sec:org34adaca}

As a turn-based-combat-enabled case-study, we ran Buddy with gpt-6-astra at high reasoning effort, using the Hermes Agent harness and the default hero personality in Godot 4.6.2. The goal prompt used was "Complete the main quest storyline." Unlike the benchmark runs in Appendix A, this run used normal turn-based combat, enabling Buddy to take damage and die if encounters went poorly. Combat decisions were made through structured game-state tools, similar to how it interacts with other parts of the environment.

The recorded trace spanned 41 minutes 8 seconds. The game engine marked the questline as completed after 39 minutes 9 seconds, following the defeat of the Goblin King. Buddy then checked its quest state, spent its remaining level-up choice, and returned to the castle before calling \texttt{complete\_goal} at 41 minutes 2 seconds. The final state lists seven completed quests and no active quests, with the character at level 6. The shutdown analytics record no player deaths.

\begin{table}[t]
\centering
\small
\caption{Combat encounters in the case-study run. Time is measured between engine-recorded combat entry and exit, in minutes and seconds. \emph{Actions} counts all \texttt{combat\_action} invocations, including calls inside generated code, failed escape attempts, and one rejected attack. HP gives player health at entry and exit.}
\label{tab:combat-case-study}
\begin{tabularx}{\columnwidth}{@{}Xrrr@{}}
\toprule
Encounter (outcome) & Time & Actions & HP \\
\midrule
Skeletons (won) & 2:01 & 15 & 106 / 51 \\
Bandits (retreated) & 2:06 & 13 & 110 / 5 \\
Bandits (won) & 2:02 & 11 & 110 / 79 \\
Goblin King (won) & 1:44 & 6 & 133 / 61 \\
\bottomrule
\end{tabularx}
\end{table}
\subsection{Retreat and Preparation}
\label{sec:orgb69413c}

The run did not proceed through combat without difficulty. Buddy first defeated three skeletons in the Abbey graveyard using basic attacks, Cleave, and a health potion. Before continuing on with the quest, it replaced its starting broken sword with a Warhammer and bought a second health potion on top of the one the player begins the game with. The most difficult encounter was a quest involving several waves of bandits. Buddy defeated the first wave, but had already consumed its potion and had 42 of 110 HP when the next wave arrived. It attempted to flee four times, escaping on the fourth attempt with only 5 HP remaining.

Rather than immediately retrying or grinding additional levels, Buddy returned to the castle to look for supplies. It bought and equipped a helmet, purchased throwable poison, and recruited two NPC allies through the hireling broker. By the time it returned to the encounter, its health had recovered to 110 HP. It was still level 4, but its reported armor had increased from 2 to 8 and it now had two allies. The second attempt cleared both waves and ended with 79 HP remaining. This is a useful example of combat changing the agent's activities outside the battle itself: the unsuccessful first attempt led to shopping, recruitment, and a return to the same objective with different preparation.

After completing the trials, Buddy equipped recovered armor, had the quest's runes bound to the Amulet of Power, and recruited the two allies again before seeking the Goblin King. In the final battle it used Fireball twice, throwable poison once, and three basic attacks. The engine recorded victory with 61 of 133 HP remaining. Across the run, the four combat entries in Table \ref{tab:combat-case-study} cover the skeleton encounter, the two bandit attempts, and the final boss; there were no additional combat encounters for experience grinding.
\subsection{Feedback and Recovery}
\label{sec:org1559717}

Combat also provides a concrete example of the tool composition described in the architecture section. Buddy made 85 \texttt{run\_code} calls during the run, and 35 of its 45 \texttt{combat\_action} calls occurred inside generated Python. These blocks selected targets, checked health, and tracked ability cooldowns from returned combat events and state changes. These changes were not all correctly handled; during the skeleton encounter, the generated code block did not remove defeated targets from the encounter, resulting in failed attempts to attack an already slain enemy. Buddy then requested fresh combat state, used a potion, and continued with a revised loop targeting the remaining enemies.

The same observation-and-correction pattern appeared during exploration. While seeking recovery after the retreat, Buddy traveled to a building it had identified as an inn. It then recognized the location as Edda's burned house and corrected the saved waypoint description before continuing its search. The run also crossed three heartbeat-initiated agent turns and one context compaction, with no additional human messages recorded. The first two turns ended with summaries that explicitly said the invasion remained unresolved; the persistent goal kept the agent working until the final quest state was verified.
\end{document}